\pdfoutput=1

\documentclass[11pt]{article}
\usepackage{EACL2023}

\usepackage{times}
\usepackage{latexsym}

\usepackage[T1]{fontenc}

\usepackage[utf8]{inputenc}

\usepackage{microtype}

\usepackage{inconsolata}

\usepackage{multirow}
\usepackage {xcolor}
\title{Preparing the Vuk'uzenzele and {ZA}-gov-multilingual South African multilingual corpora}

\author{Richard Lastrucci\textsuperscript{1}, Isheanesu Dzingirai\textsuperscript{1}, Jenalea Rajab\textsuperscript{2}, Andani Madodonga\textsuperscript{1}, \\
     {\bf Matimba Shingange}\textsuperscript{1}, {\bf Daniel Njini}\textsuperscript{1}, {\bf Vukosi Marivate}\textsuperscript{1,3} \\
    \textsuperscript{1}Department of Computer Science, University of Pretoria\\
    \textsuperscript{2}School of Computer Science and Applied Mathematics, University of the Witwatersrand\\ 
    \textsuperscript{3}Lelapa AI \\
    \texttt{richard.lastrucci@tuks.co.za}, \texttt{ishe.dzingirai@gmail.com}, \texttt{jenalea.rajab@gmail.com}, \\
    \texttt{andanim412@gmail.com}, \texttt{mrosslyns@gmail.com}, \texttt{vukosi.marivate@cs.up.ac.za}
  }

\begin{document}
\maketitle
\begin{abstract}
This paper introduces two multilingual government themed corpora in various South African languages. The corpora were collected by gathering the South African Government newspaper (Vuk'uzenzele), as well as South African government speeches (ZA-gov-multilingual), that are translated into all 11 South African official languages. The corpora can be used for a myriad of downstream NLP tasks. The corpora were created to allow researchers to study the language used in South African government publications, with a focus on understanding how South African government officials communicate with their constituents. 

In this paper we highlight the process of gathering, cleaning and making available the corpora. We create parallel sentence corpora for Neural Machine Translation (NMT) tasks using Language-Agnostic Sentence Representations (LASER) embeddings. With these aligned sentences we then provide NMT benchmarks for 9 indigenous languages by fine-tuning a massively multilingual pre-trained language model.  
\end{abstract}

\section{Introduction}

The advancement of Natural Language Processing (NLP) research in Africa is impeded due to the scarcity of data for training models for various NLP tasks \cite{nekoto2020participatory} as well as availability of benchmarks and ways to reproduce them \cite{martinus2019focus}. For many South African languages there are still challenges finding easily available textual datasets \cite{marivate2020investigating} even if there are many speakers for those languages \cite{ranathunga-de-silva-2022-languages}. There is a need to focus on development of local language \cite{joshi-etal-2020-state} NLP resources.

This paper builds upon the work of Autshumato \cite{groenewald2009introducing,groenewald2010processing} by creating automatically aligned parallel corpora from government textual data in the 11 official languages of South Africa. While the Autshumato project focused on creating Machine Translation tools for five indigenous languages, the resulting corpora lacked information about its origin or context, limiting its usefulness for other NLP tasks such as categorisation, topic modelling over time and other tasks that require contextual information of the content. Our approach provides more comprehensive data that can support a wider range of NLP applications.

Our belief is that there is a significant opportunity to create a more user-friendly data collection process that can be easily maintained and provide extraction tools for others. It is essential to preserve the data source and structure it in a way that enables extensions. Our goal is to enhance Neural Machine Translation (NMT) resources in the government data domain by including all indigenous languages and broadening the translation directions beyond English as the source language. Additionally, we recognise the importance of providing aligned data across all South African languages beyond English.

Further, this paper introduces parallel corpora datasets in the 11 official languages of South Africa, created from text data obtained from the government. These datasets are designed to facilitate the development of NMT models. The corpora are automatically aligned, and are expected to serve as a valuable resource for researchers and practitioners working in the field of machine learning.

The parallel corpora 
were generated using LASER encoders \cite{schwenk2017learning}, facilitating the one-to-one alignment of tokenised sentence data. The data was sourced from credible sources such as newspapers and academic journals and covers 
diverse topics including health, finance, and politics.

We also provide NMT benchmarks for the parallel corpora by fine-tuning a massively multilingual model (M2M100 \cite{fan2021beyond}) building on the work of \citet{lafand}.

This paper is structured as follows. In the following section, we detail the datasets that we have compiled, including their compilation methodology and the information they contain. We then describe how we have aligned and created parallel corpora using these datasets. The subsequent section presents our NMT experiments and provides an analysis of the results obtained. Finally, we conclude the paper with our findings and make recommendations for future research.

\section{Main Datasets}
\subsection{The Vuk'uzenzele South African Multilingual Corpus}

The  Vuk'uzenzele dataset was constructed from editions of the South African government magazine Vuk'uzenzele\footnote{\url{https://www.vukuzenzele.gov.za/}}. Being a magazine, the text focuses mainly on current events, politics, entertainment, and other topics related to a magazine publication. The Vuk'uzenzele dataset provides a comprehensive view of the language and topics of discussion in South Africa during the respective period, giving researchers insight into the history and culture of South Africa. The  Vuk'uzenzele dataset is thus a rich resource for any researcher wanting to analyse South African politics, current events, and popular culture. 

\subsubsection{Creation of Vuk'uzenzele}

The raw Vuk'uzenzele data is scraped from PDF editions of the magazine. The main Vuk'uzenzele edition is in the English language. Only a few of these English articles are translated into the other 10 official South African languages (Afrikaans, isiNdebele, isiXhosa, IsiZulu, Sepedi, Sesotho, siSwati, Tshivenda, Xitsonga and Setswana).  As such, we created a pipeline to identify which articles should be extracted from each language pdf from a specific edition. Individual articles were extracted and placed into text files. The extracted text files still have some challenges due to PDF extraction. To clean it, a team member goes through each extracted text file and formats it as follows: 
\begin{itemize}
    \item Line 1: Title of article (\emph{in language})
    \item Line 2: \emph{empty line}
    \item Line 3: Author of article (\emph{if available. If not, defaults to Vukuzenzele Unnamed}) 
    \item Line 4: \emph{empty line}
    \item Line 5-end: \emph{body of article}
\end{itemize}

The data is easier to analyse and visualise after being manually reviewed. The labour-intensive effort was necessary to provide a comprehensive and meaningful analysis of the magazine’s content. The time-consuming process of manual review and extraction was ultimately worth it, as it provides an opportunity to create a deeper understanding of the content within Vuk'uzenzele. As of writing we have \emph{53} editions of the newspaper spanning \emph{January 2020 to July 2022}. More additions will be added in time by the team. Automations have been built to download and archive the PDFs, however manual effort is still required to extract and identify translated articles. The dataset, code and automated scrapers are available at at \url{https://github.com/dsfsi/vukuzenzele-nlp} and Zenodo\footnote{\url{https://doi.org/10.5281/zenodo.7598539}} \cite{marivate_vukuzenzele}. We make it available in a format that allows other researchers to extend, remix and add onto it (\emph{CC-4.0-BY-SA licence for data and MIT License for code}).  

\subsection{The ZA-Gov Multilingual South African corpus}
The ZA-Gov Multilingual corpus dataset was constructed from the speeches following cabinet meetings of the SA government. As such, the dataset carries a variety of topics including energy, labour, service delivery, crime, COVID, international relations, the environment, and government affairs such as government appointments, cabinet decisions, etc. This provides an eye into the workings of the South African government and how it has dealt with various challenges, both internal and external. 

\subsubsection{Creation of ZA-Gov-multilingual}
The raw ZA-Gov Multilingual data is scraped from the the South African government website (\url{https://www.gov.za/}), where all cabinet statements, and their translations, are posted. The data was extracted and structured into a JSON format. The JSON payload for each speech records:

\begin{itemize}
  \item Date,
  \item Datetime,
  \item Title (\emph{in English}),
  \item Url (\emph{top url for speech}), 
  \item Language payload for each language (\emph{eng, afr, nbl, xho, zul, nso, sep, tsn, ssw, ven, tso}).
  \begin{itemize}
     \item Title (\emph{in language}),
     \item Text (\emph{in language}),
     \item Url (\emph{for the translation}).
  \end{itemize}
\end{itemize}

This structure makes it convenient for researchers and analysts to perform various natural language processing, data mining and machine learning tasks such as sentiment analysis, topic modelling, categorisation, language modelling and more. For instance, through sentiment analysis and text mining, analysts can investigate opinions of cabinet members' statements and track the evolution of these topics over time. As of writing, the dataset contains \emph{162} cabinet statements spanning \emph{2 May 2013 to 1 December 2022}. The dataset will update automatically when new, \emph{translated}, statements are available on the gov.za website. The dataset, code and automated scrapers are available at at \url{https://github.com/dsfsi/gov-za-multilingual} and Zenodo\footnote{\url{https://doi.org/10.5281/zenodo.7635167}}
\cite{marivate_gov_za}. We make it available in such a way that other researchers can extend, remix and add onto it (\emph{CC-4.0-BY-SA licence for data and MIT License for code}).  

\section{The corpora as a foundation for other NLP tasks and further study}

In addition to supporting the creation of NMT models (discussed in the proceeding section), our datasets have the potential to serve as a foundation for many other NLP tasks beyond translation. We believe that these datasets will be a valuable resource for the study of South African government communication, and that it can be used for direct creation of multilingual document categorisation/classification \cite{schwenk2018corpus}, simplification \cite{lu2021unsupervised,siddharthan2014survey, martin2022muss}, entity extraction \cite{tedeschi-etal-2021-wikineural-combined,chen2018cotraining,pappu2017lightweight,emelyanov2019multilingual}, and other NLP tasks. To further extend the dataset's usefulness, we recommend looking at work such as the Parallel Meaning Bank \cite{abzianidze2017parallel}, which can act as an inspiration for transferring knowledge from one language to another and provide new benchmarks that may be helpful for Southern African languages beyond South Africa. We envision these datasets as a starting point for further research in the area of multilingual NLP for South African and African languages.


\section{Methods for Processing and Compilation}

The datasets are a two way parallel corpus of the 11 official languages of South Africa, which are listed in Table~\ref{table1} with their corresponding ISO 639-2code. The datasets contain texts written in the official languages of South Africa, including Afrikaans, English, isiNdebele, isiXhosa, isiZulu, siSwati, Sepedi, Xitsonga, siSwati, Tshivenda, and Setswana. As such, there are 55 ways of combining these languages into pairs, producing 55 distinct corpora in each of the datasets.  The dataset uses the ISO 639-2 language codes in its naming convention, i.e., 'aligned-afr-zul.csv'. By nature of compilation, some datasets have more observations than others, which could lead to varying results, i.e., if used for NMT, then a better model can be produced for two languages from a dataset with more observations as opposed to one with fewer observations.  This compilation of data allows for further exploration into the complexities of South African language and discourse, creating a multi-dimensional representation of how language is used and interpreted in South Africa. Through these datasets, the range of language usage in South Africa can be explored, providing insights into how different languages interact.

\begin{table}[ht!]
\centering
\caption{Language List with ISO 639-2 codes}
\label{table1}
\begin{tabular}{|p{2cm}|p{2cm}|}
\hline \hline
 Name &  Code \\
\hline \hline
 isiZulu   & zul \\
 isiXhosa  & xho \\
 Afrikaans & afr \\
 English   & eng \\
 Sepedi    & nso \\
 Setswana  & tsn \\ 
 Xitsonga &  tso \\
 Sesotho & sot \\
 siSwati &  ssw \\
 Tshivenda &  ven \\
 isiNdebele & nbl \\
 \hline
\end{tabular}
\end{table}

\subsection{Preprocessing}
Preprocessing was required to refine the raw scraped data prior to LASER encoding and alignment. The preprocessing steps are listed below and differ slightly as the source data and method of scraping has an outcome on the data obtained. For example, ZA-Gov-Multilinguals involve a lot of nested points, i.e., \textit{2.1.2}, which needed to be removed, while in contrast the Vuk`uzenzele data uses bullet points for listing. 

\subsubsection{Vuk'uzenzele}
The raw text from the collected data was pre-processed in the following way:
\begin{itemize}
\item The text was set to lowercase.
\item Hyphens and bullet points were removed. 
\item Double spaces, tabs, and newlines were replaced with a single space.
\item The standard apostrophe, i.e., '', took the place of Unicode apostrophes. 
\end{itemize}

\subsubsection{ZA-Gov-multilingual}
The raw text from the collected data was pre-processed in the following way:
\begin{itemize}
    \item Removing the dots (or single- or multi-digit numbers) that began a line
    \item Inserting a period after a series of numbers in the format $x \cdot y$. 
    \item Adding a period after a string of numbers in the format $x$. 
    \item Replacing a sequence of punctuation marks, such as a period, colon, semi-colon, or a combination of these, followed by a letter with a single period.
\end{itemize}

\subsection{Corpora Alignment}
Once preprocessed, the text was passed to the NLTK tokeniser "punkt" which returns a vector of sentence tokens.
The \textit{n} tokenised sentences were sent to LASER encoder which encodes it into \textit{n} sentence vectors, each of length 1024. The sentence vectors are compared and a cosine similarity algorithm was performed to produce a score from 0 to 1 on the similarity of the two vectors as described in section \ref{LASER}.

\subsubsection{LASER Encodings}
\label{LASER}
In order to compare the text for similarity, LASER encoders were utilised. LASER, which stands for Language-Agnostic Sentence Representations, is a research project by Facebook AI Research. LASER generates sentence representations by encoding sentences into a vector. The vectors serve as a machine representation of the sentence. The vectors can be compared using cosine similarity which outputs a score between zero and one. Cosine scores closer to one indicate high similarity. This score is recorded in the LASER datasets (available in the dataset repository). The observations present in each dataset with a score above 0.65, or 65\% similarity, are listed in the following tables \ref{table2} and \ref{table3}.
Entire tables featuring the number of observations present in all datasets are featured on the READMEs of the dataset repos, \url{https://github.com/dsfsi/gov-za-multilingual} for ZA-Gov-Multilingual and \url{https://github.com/dsfsi/vukuzenzele-nlp} for  Vuk'uzenzele. 

\begin{table}[ht!]
\centering
\caption{Top ten datasets with the most observations with a cosine score greater than or equal to 0.65 in Vuk'uzenzele.}
\label{table2}
\begin{tabular}{|p{2cm}|p{4cm}|p{3cm}|}
\hline \hline
Language pair & No. of observations in Vuk'uzenzele \\
\hline \hline
ssw-xho & 2,202 \\
ssw-zul & 2,183 \\
xho-zul & 2,102 \\
nso-xho & 2,081 \\
nso-tso & 2,071 \\
ssw-tso & 2,034 \\
nso-ssw & 2,021 \\
tsn-tso & 2,020 \\
tsn-xho & 2,009 \\
tso-xho & 2,009 \\
 \hline
\end{tabular}
\end{table}

\begin{table}[ht!]
\centering
\caption{Top ten datasets with the most observations with a cosine score greater than or equal to 0.65}
\label{table3}
\begin{tabular}{|p{2cm}|p{3cm}|p{3cm}|}
\hline \hline
Language pair & No. of observations in ZA\-gov Multilingual \\
\hline \hline
nbl-ven & 18,984 \\ 
nso-ssw & 18,697 \\ 
zul-ssw & 18,563 \\ 
xho-ssw & 18,387 \\ 
xho-zul & 18,145 \\ 
xho-nso & 18,110 \\ 
xho-tso & 17,954 \\ 
ssw-tso & 17,880 \\ 
zul-tso & 17,789 \\ 
zul-nso & 17,630 \\ 
 \hline
\end{tabular}
\end{table}

\subsection{Postprocessing}
For the LASER datasets the source sentence, target sentence, and cosine score for the aligned data was written to a csv file with the naming convention 'aligned-\{src\_lang\_code\}-\{tgt\_lang\_code\}.csv', i.e. 'aligned-afr-zul.csv'. Refer to the language list in \ref{table1} for language codes used in naming the datasets.

For the simple aligned datasets the source sentence and the target sentence were written to a csv file with the same naming structure as the LASER datasets.

\section{NMT Benchmarks}
Minimal aligned sentence corpora, for low-resourced African languages, hinder the quality of NMT models trained from scratch \cite{martinus2019focus, nekoto2020participatory,lafand}. Recently \citet{lafand} approached this problem by fine-tuning massively multilingual models, including the M2M100 model \cite{fan2021beyond}, on a small number of aligned sentences. The M2M100 model is a Many-to-Many non-English centric language model trained to translate directly between 100 languages, including five South African official languages \cite{fan2021beyond}. \citet{lafand} demonstrated how to effectively leverage this model for small quantities of data, to create NMT systems for languages and domains not included in pre-training. 

Building on their work, we create baseline translation benchmarks for the Vuk’uzenzele and ZA-gov-multilingual datasets, in the government publication domain, by fine-tuning the M2M100 model. To provide our results in context and for comparison purposes we also fine-tune the M2M100 model on subsets of the existing Autshumato parallel corpora obtained from the South African Centre for Digital Language Resources \cite{isiNdebele, Sesotho, Setswana, Xitsonga, Tshivenda, IsiZulu} 
(\url{https://sadilar.org}). We focus our efforts on providing NMT benchmarks for the low resource African languages in the datasets, as such Afrikaans translations are not included due to the relatively high availability of digital datasets in this language, and we leave this for future work.

\subsection{Pre-processing}
The aligned datasets were processed to remove duplicate and conflicting translations (in both the source and target sentences) then shuffled to remove any order bias before the train, test and dev set were created. The data splits are defined as 70\% training, 20\% test and 10\% dev sets. For comparison, all models are fine-tuned using the `xxx-eng' translation direction where `xxx' represents the indigenous African source language and `eng' is the English translation target. 

The available Aushumato parallel corpora (extracted from various government resources and web-crawls \cite{groenewald2009introducing}) are comparable in domain to the ZA-gov-multilingual parallel corpora created, however the dataset sizes are currently much larger. We therefore extract the same number for pre-processed aligned sentences as the ZA-gov-multilingual corpora in the `xxx-eng' translation direction, for direct NMT result comparison. The sentence and token counts of the corpora used for NMT benchmarking are provided in Appendix \ref{app:Data Statistics1} tables \ref{DataTable1}, \ref{DataTable2} and \ref{DataTable3}.

\subsection{Results}
The M2M100 fine-tuning benchmark results for the  Vuk’uzenzele, ZA-gov-multilingual and subsets of the available Autshumato parallel corpora are provided in table \ref{NMTTable}. The fine-tuning translation directions are provided, and any source languages which were not including in the original M2M100 pre-training are highlighted for references purposes. Additionally the highest BLEU score result achieved across the datasets is shown in bold. Cases where the Autshumato parallel corpora were not accessible or did not exist for a particular language are shown with a `-' symbol.

\begin{table}[ht!]
\centering
\caption{BLEU scores for Massively Multilingual Transfer on xxx-eng translations using the Vuk’uzenzele (Vuk.), ZA-gov-multilingual (Gov.) and subsets of the available Autshumato datasets (Aut.)}
\label{NMTTable}
\begin{tabular}{|l|l|l|l|}
\hline \hline

\multirow{2}{2cm}{\centering Translation Direction}  &  \multicolumn{3}{c|}{BLEU} \\  
\cline{2-4} 
   &  Vuk. & Gov. & Aut.\\
\hline \hline
  \color{black}\textbf{\texttt{nbl}$\rightarrow$\texttt{eng}}  &7.33 & 8.04 & \textbf{12.24}\\
 \texttt{nso}$\rightarrow$\texttt{eng} &9.29 & \textbf{26.50} & - \\
 \texttt{ssw}$\rightarrow$\texttt{eng} &4.80 & \textbf{28.72} & - \\
 \color{black}\textbf{\texttt{sot}$\rightarrow$\texttt{eng}} &4.55 & 10.21 & \textbf{14.83}\\
 \texttt{tsn}$\rightarrow$\texttt{eng} &2.80 &  \textbf{29.68} & 28.04 \\
  \color{black}\textbf{\texttt{tso}$\rightarrow$\texttt{eng}} &13.86 & \textbf{35.40} & 32.10\\ 
 \color{black} \textbf{\texttt{ven}$\rightarrow$\texttt{eng}} &2.32 & 9.68 &  \textbf{17.24}\\
 \texttt{xho}$\rightarrow$\texttt{eng} &6.05 & 26.81& - \\
 \texttt{zul}$\rightarrow$\texttt{eng} &9.97 & \textbf{30.03} & 25.90\\
 \hline
\end{tabular}
\end{table}

Fine-tuning on the Vuk’uzenzele datasets achieved the lowest overall BLEU scores, this is expected due to the small size of the aligned datasets in comparison to those of the ZA-gov-multilingual and Autshumato datasets. The highest BLEU scores are distributed inconsistently across the ZA-gov-multilingual and Autshumato NMT models, with the ZA-gov-multilingual models achieving a higher score for Setswana, Xitsonga and isiZulu. This could be due to the random subset selections from the Autshumato datasets, as well as a result of variations in source combinations and cleaning methods used by the Autshumato project when creating the aligned corpora. It is noted that variations in the subset selections will yield different results and an in-depth analysis is left for future work. 

We achieve the highest benchmark result for Xitsonga (`tso-eng') across all datasets, which is a language that the M2M100 model has not been pre-trained on, demonstrating the effectiveness of transfer learning for new low-resource language datasets.

It is also noted that current NMT resources for the Autshumato datasets exist only in the `eng-xxx' translation direction for Xitsonga, Setswana, Sepedi, Sesotho and isiZulu \cite{LIT1766}. Our contributions therefore extend the benchmark translation resources (in the government data domain) to isiNdebele, isiXhosa, siSwati and Tshivenda; and broaden the translation direction beyond English as the source language.

\section{Conclusion}

Finally, this paper presented two multilingual corpora that are automatically aligned to facilitate the translation of texts between languages. These datasets contribute to an expanding collection of corpora for training African language NMT models that are left out or under-resourced in contemporary NLP research. It is the hope of the authors that these datasets will aid in creating NMT models for low-resource African languages and that the models can be used to facilitate access to translation services, empowering African speakers and writers to communicate more effectively in their native languages. It is also hoped that the datasets will further NLP research into the multilingualism of African languages and contribute to an understanding of the various dialects present in Africa.

\section{Limitations}

The NMT models discussed have been created for benchmarking the described datatsets and have not been exhaustively quality tested for production purposes. We also only tested the effectiveness of fine-tuning the M2M100 model with English as a target language and have have not extended the NMT systems to translations between indigenous South African languages, therefore further benchmarking still needs to be implemented. We would like to extend testing to include the evaluation set created by \cite{McKellar:2020aa}, which contains data (excluded from the Autshumato corpora) for all 11 official South African languages and could provide a more accurate comparison of the NMT models. It is noted that while the BLEU score results are promising, a qualitative linguistic analysis still needs to done on the translation models to determine if the BLEU scores for certain language translations (i.e. 'tso-eng') correlate to accurate translations within our domain. As future work we hope to collaborate with respective linguists to improve the quality and effectiveness of such NMT systems for South African Languages \cite{LIT1766}. 

\section{Ethics Statement}
The datasets created and used for the translation model benchmarks are taken solely from South African government resources. Therefore it is highlighted that if these models are used in production, they might ignore certain social/societal structures and will be representative of the dominant political party at the time the datasets were sourced \cite{parrots}. We also note that the benchmark models and datasets have not been curated to determine any biases that are present. As such, any existing biases in the system might have the potential to harm specific groups, when used in NLP downstream production tasks \cite{parrots}.

\section{Acknowledgements}

We would like to acknowledge funding from the ABSA Chair of Data Science, the Google Research scholar program, TensorFlow Award for Machine Learning Grant and the NVIDIA Corporation hardware Grant. Many thanks to all the anonymous RAIL reviewers for their valuable feedback.

\bibliography{custom}
\bibliographystyle{acl_natbib}

\newpage
\appendix

\section{Appendix}
\label{Appendix}
\subsection{Data Statistics}\label{app:Data Statistics1}

The data statistics for the datasets used for NMT bench-marking are provided in Tables \ref{DataTable1}, \ref{DataTable2} and \ref{DataTable3}.

\begin{table}[h]
\centering
\caption{ Characteristics of the translation data for the Vuk’uzenzele (Vuk.) datasets}
\label{DataTable1}
\begin{tabular}{|l|l|}
\hline \hline

\multirow{2}{2cm}{\centering \textbf{Translation Direction}}  & \multirow{2}{5cm} {\centering \textbf{Size} \\ \textbf{\#sents (\#src / \#trg tokens)}} \\ &\\

\hline \hline
 \texttt{nbl}$\rightarrow$\texttt{eng} & 136 (3.4k / 3.9k)\\
 \texttt{nso}$\rightarrow$\texttt{eng} & 1715 (53.7k / 41.6k)\\
 \texttt{ssw}$\rightarrow$\texttt{eng} & 1588 (29.9k / 37.6k)\\
 \texttt{sot}$\rightarrow$\texttt{eng} & 260 (9.9k / 7.5k)\\
 \texttt{tsn}$\rightarrow$\texttt{eng} & 1366 (49.8k / 31.7k)\\
 \texttt{tso}$\rightarrow$\texttt{eng} & 1998 (58.9k/ 46.6k)\\ 
 \texttt{ven}$\rightarrow$\texttt{eng} & 230 (9.1k / 7k)\\
 \texttt{xho}$\rightarrow$\texttt{eng} & 1338 (25.8k / 31.5k)\\
 \texttt{zul}$\rightarrow$\texttt{eng} & 1874 (34.1k / 43k)\\
 \hline
\end{tabular}
\end{table}

\begin{table}[h]
\centering
\caption{ Characteristics of the translation data for the ZA-gov-multilingual (Gov.) datasets}
\label{DataTable2}
\begin{tabular}{|l|l|}
\hline \hline

\multirow{2}{2cm}{\centering \textbf{Translation Direction}}  & \multirow{2}{5cm} {\centering \textbf{Size} \\ \textbf{\#sents (\#src / \#trg tokens)}} \\ &\\
\hline \hline
 \texttt{nbl}$\rightarrow$\texttt{eng} & 3513 (63.9k / 107k)\\
 \texttt{nso}$\rightarrow$\texttt{eng} & 14742 (460.9k / 375k)\\
 \texttt{ssw}$\rightarrow$\texttt{eng} & 15139 (291k/ 377.8k)\\
 \texttt{sot}$\rightarrow$\texttt{eng} & 4995 (145.9k / 153.5k)\\
 \texttt{tsn}$\rightarrow$\texttt{eng} & 14068 (493.1k / 362.2k)\\
 \texttt{tso}$\rightarrow$\texttt{eng} & 15393 (466.4k / 381.2k)\\ 
 \texttt{ven}$\rightarrow$\texttt{eng} & 3404 (68.2k / 96.6k)\\
 \texttt{xho}$\rightarrow$\texttt{eng} & 15853 (318.2k / 389.5k)\\
 \texttt{zul}$\rightarrow$\texttt{eng} & 15503 (327.5k / 384.1k)\\
 \hline
\end{tabular}
\end{table}

\begin{table}[h!]
\centering
\caption{ Characteristics of the translation data for the subsets of the
available Autshumato datasets (Aut.)}
\label{DataTable3}
\begin{tabular}{|l|l|}
\hline \hline

\multirow{2}{2cm}{\centering \textbf{Translation Direction}}  & \multirow{2}{5cm} {\centering \textbf{Size} \\ \textbf{\#sents (\#src / \#trg tokens)}} \\ &\\
\hline \hline
 \texttt{nbl}$\rightarrow$\texttt{eng} & 3513 (48.4k / 65k)\\
 \texttt{sot}$\rightarrow$\texttt{eng} & 4995 (118.3k / 100.4k)\\
 \texttt{tsn}$\rightarrow$\texttt{eng} & 14068 (335.1k / 274.7k)\\
 \texttt{tso}$\rightarrow$\texttt{eng} & 15393 (193.7k/ 166k)\\ 
 \texttt{ven}$\rightarrow$\texttt{eng} & 3404 (80.6k / 65.9k)\\
 \texttt{zul}$\rightarrow$\texttt{eng} & 15503 (231.8k / 307.6k)\\
 \hline
\end{tabular}
\end{table}

\end{document}